\RequirePackage{fix-cm}
\documentclass[smallcondensed]{svjour3}       % onecolumn (second format)
\usepackage{makecell}
\usepackage{color,soul,xcolor}
\usepackage[paper=a4paper]{geometry}
\usepackage[utf8]{inputenc}
\usepackage{graphicx, amsmath, amssymb,algorithm,algpseudocode}

\usepackage{makecell}
\usepackage{multirow}

\begin{document}
\title{Improved Binary Artificial Bee Colony Algorithm}
\author{Rafet Durgut}
\institute{R. Durgut \at
              Karabuk University, Engineering Faculty, Computer Engineering Department, Karabuk, Turkey \\
              Tel.: +90-370-418-70-50\\
              Fax: +90-370-418-70-01\\
              \email{rafetdurgut@karabuk.edu.tr}
}
\date{Received: date / Accepted: date}
\maketitle

\begin{abstract}
The Artificial Bee Colony (ABC) algorithm is an evolutionary optimization algorithm based on swarm intelligence and inspired by the honey bees' food search behavior. Since the ABC algorithm has been developed to achieve optimal solutions by searching in the continuous search space, modification is required to apply this method to binary optimization problems. In this paper, we improve the ABC algorithm to solve binary optimization problems and call it the improved binary Artificial Bee Colony (ibinABC). The proposed method consists of an update mechanism based on fitness values and processing different number of decision variables. Thus, we aim to prevent the ABC algorithm from getting stuck in a local minimum by increasing its exploration ability. We compare the ibinABC algorithm with three variants of the ABC and other meta-heuristic algorithms in the literature. For comparison, we use the well-known OR-Library dataset containing 15 problem instances prepared for the uncapacitated facility location problem. Computational results show that the proposed method is superior to other methods in terms of convergence speed and robustness. \textit{The source code of the algorithm will be available on GitHub after reviewing process.}
\end{abstract}

\section{Introduction}
In recent years, several meta-heuristic optimization algorithms that are influenced by various phenomena of nature have been developed \cite{ref1}. There are many algorithms inspired by physical, chemical or biological phenomena and swarms of animals. An optimization algorithm is called population-based if it searches the best solution using a set of solutions \cite{ref2}. Population-based methods are divided into two parts as evolutionary algorithms and swarm intelligence algorithms. Genetic Algorithm (GA) \cite{ref3}, Evolutionary Strategy (ES)\cite{ref4}, Differential Evolution (DE)\cite{ref5} are the most popular in population-based evolutionary algorithms, while Artificial Bee Colony (ABC)\cite{ref6}, Particle Swarm Optimization (PSO)\cite{ref7}, Grey Wolf Optimization (GWO)\cite{ref8}, Crow Search Algorithm (CSA)\cite{ref9} and Whale Optimization Algorithm (WOA)\cite{ref10} are the most popular in population-based swarm intelligence algorithms.

Optimization methods work on certain types of problems when they are first proposed. These optimization problems can be continuous (ABC, PSO), combinatorial (GA), binary (GA) or constrained under some conditions (CSA). An optimization algorithm can be applied to other problem types. However, by the NFL theorem \cite{ref11}, the increase in success for one problem type has the opposite effect in other problem types. Therefore, when an algorithm is applied to different types of problems, some improvements are required. The literature on the variants of meta-heuristic algorithms developed for different problems is quite extensive \cite{ref12} \cite{ref13}.

The ABC algorithm was first developed by applying to continuous problems. Multiobjective \cite{ref14}, Binary \cite{ref15}, Combinatorial \cite{ref16} versions are available in the literature. Although the ABC algorithm can easily evolve optimization problems, it has some structural problems. Exploration-exploitation balance is not provided properly with existing operators and mechanisms. In the original ABC algorithm, the exploration phase supported by many mechanisms remained more dominant and the exploitation phase remained more recessive. Only one decision variable is updated at each iteration in the original ABC algorithm. This situation differs for binary versions. The discrete binary ABC (DisABC) algorithm which is a similarity-based binary variant proposed by Kashan et al. generates a new candidate solution by changing the value of more than one bit \cite{ref17}. Although the DisABC algorithm provides fast convergence by changing several decision variables, it works slowly in large-scale problems due to the fact that it contains the mixed integer linear programming model. Kıran et al. proposed a new variant of the ABC algorithm based on XOR operations.  A candidate solution is generated by a mechanism derived in a similar way to the original ABC update mechanism \cite{ref18}. However, Kıran et al. adjust the balance of exploration and exploitation with a parameter in the mechanism. But, the effect of this parameter has not been revealed in their study.

Solving binary problems is very substantial in many areas, in particular  computer, mathematics and economics. \cite{binaryreview}. Therefore, there are also arrangements of different meta-heuristics proposed to solve binary  optimization problems in the literature \cite{binarywork}. Chuang et al. improved the binary PSO for solving the feature selection problem \cite{psobinary}. Kiran et al. made a modification of the artificial algae algorithm that is improved for continuous optimization problems to solve Uncapacitated Facility Location Problem (UFLP) \cite{ref22}.

This paper consists of an improved version of the binABC algorithm proposed by Kıran et al. \cite{ref18} by enhancing the ability of exploitation. This paper aims to obtain better results in solving binary optimization problems using the ABC algorithm. With this purpose, we propose an improved update mechanism. The proposed mechanism consists of two parts. The proposed mechanism includes two parts. The first is the determination of the number of decision variables to be changed in each iteration as non-linear and stochastic, which strengthens the exploration at the early phase and the exploitation in the middle and the last phase. The second is the use of an improved update operator to transfer better solutions to the next generations.

 The remainder of the paper is organized as follows. In Section 2, the working principle of the Artificial Bee Colony Algorithm is explained. The studies which  apply the ABC algorithm to binary problems are presented in Section 3. In Section 4, the enhancements proposed in this paper are given. In Section 5, the Uncapacitated Facility Location Problem which is used to test the proposed algorithm is introduced. Experimental studies are carried out to obtain the best parameter configuration of the  proposed method in Section 5. In Section 6, an extensive comparison of the proposed method with the other methods in the literature is given and finally, some concluding remarks are given in Section 7. 

\section{Artificial Bee Colony Algorithm}
The artificial bee colony algorithm was developed by Karaboğa inspired by the honey bees' food search behavior. The algorithm was first applied to continuous optimization problems \cite{ref6}. The ABC algorithm models the swarm intelligence formed by bees interacting with each other in the honey hive. According to the model, there are three types of bees in the hive; employed bees, onlooker bees and scout bees. These bee types are modeled in a way that each of them performs a phase. In the first phase, the so-called employed bee phase, each employed bee tries to improve its own food source. In the onlooker bee phase, each onlooker works on its own food source in proportion to the quality of its food source. In the scout bee phase, if the onlooker bees fail to provide an improvement in the food source, the scout bees start the search for a new food source.

A food source represents a feasible solution in the model. The fitness function, which depends on the cost function of this solution, expresses nectar information and is calculated by Equation\ref{eq1}. Depending on the probability value calculated by Equation\ref{eq2} according to this fitness value, the neighborhood operator is applied to the current food source as in Equation \ref{eq3}. The limit value determined for the implementation of the scout bee phase is controlled by a variable called trial and if this value exceeds the limit value, a new solution is generated by Equation \ref{eq4}.

\begin{equation}
\label{eq1}
Fit(\textbf{x}_i)  = \begin{cases}
        \frac{1}{1+f(\textbf{x}_i)} & f(\textbf{x}_i)\geq 0\\
        	&   \\
        1 + |f(\textbf{x}_i)| & \text{otherwise}
        \end{cases}
\end{equation}

\begin{equation}
\label{eq2}
    p_i  = \frac{Fit(\textbf{x}_i)}{\sum_{j=1}^{N} Fit(\textbf{x}_j)}
\end{equation}

\begin{equation}
\label{eq3}
    \textbf{v}_{i,j}  = \textbf{x}_{i,j} + \textbf{$\phi$}_{i,j} (\textbf{x}_{i,j} - \textbf{x}_{n,j})
\end{equation}
\begin{equation}
\label{eq4}
    x_{i,j}=LB_j+rand(0,1)\times(UB_j-LB_j)
\end{equation}
where $x_i,x_n,v_i$ are the selected solution, neighbor solution, candidate solution, respectively. $i=1,2 ... ,N$ represents the index of the food source, where $N$ is the number of  food sources.  $j=1,2,..,D$ is the decision variable index in the $D$-dimensional solution set. $LB, UB$ represent the lower and upper bound values for a decision variable, respectively. There are many studies on a detailed analysis of the method and its applications \cite{ref19}. 
\section{Binary Artificial Bee Colony}
Since the ABC algorithm is developed for the continuous decision variables, it is not possible to apply it to binary optimization problems without making an arrangement. Therefore, the binary ABC variants have been proposed to solve this problem. Researchers improved new variants by making some arrangements on Equations \ref{eq3} and \ref{eq4}.  In what follows, we review three current studies. The methods generate new random solutions using  Equation\ref{eq5} instead of Equation \ref{eq4} by a Bernoulli process. 
\begin{equation}
\label{eq5}
    x_{i,d}=\begin{cases}
			0, & \text{if $rand<$ 0.5}\\
            1, & \text{otherwise}
		 \end{cases}
\end{equation}
\subsection{binABC}
Kıran et al. updated Equation \ref{eq4} with Equation \ref{eq6} that uses the XOR logical operator to produce a candidate solution in their binary variant \cite{ref18}. Here, parameter $\vartheta$ is used as a logic NOT gate. If it is less than the threshold value (0.5), then the output of parenthesis is complemented.
\begin{equation}\label{eq6}
   v_{i,j}=x_{i,j}\oplus\vartheta(x_{i,j}\oplus x_{n,j}) 
\end{equation}
where $i$ is selected solution, $j$ is randomly selected problem dimension. To determine the corresponding bit of the candidate solution, the corresponding bit of the selected solution and the corresponding bit of the neighbor solution are taken to a logical process. Possible candidate bit values that can be obtained using the neighborhood operator are given in Table \ref{table1}. As can be seen in Table \ref{table1}, for state 1 ($ \vartheta <0.5$), if the input bits are the same, then the output is inverted, otherwise it follows the input. For  State 2 ($ \vartheta \geq 0.5 $), if the input bits are the same, then the output takes the same value of the input, if it is different, then it takes the value of the neighbor bit.

\begin{table}[h]
\label{table1}
\caption{Truth table for XOR-based neighborhood operation}
\centering
\begin{tabular}{lllllll}
~                                   & ~                 & ~                        & state 1                      & state 2            & for state 1       & for state 2        \\ 
\hline
 $x_{i,j}$                      &  $x_{n,j}$    &  $x_{i,j} \oplus x_{n,j}$ &  $\vartheta < 0.5 $    &  $\vartheta \geq 0.5 $    &  $v_{i,j}$     &  $v_{i,j}$      \\ 
\hline
0                                   & 0                 & 0                        & 1                            & 0                  & 1                 & 0                  \\
0                                   & 1                 & 1                        & 0                            & 1                  & 0                 & 1                  \\
1                                   & 0                 & 1                        & 0                            & 1                  & 1                 & 0                  \\
1                                   & 1                 & 0                        & 1                            & 0                  & 0                 & 1                 
\end{tabular}
\end{table}

\subsection{disABC}
Kashan et al. proposed a dissimilarity based ABC (disABC) variant for binary optimization problems \cite{ref17}. The disABC method uses a new solution generator by transforming equation \ref{eq4} into the binary search space. The new solution generator calculates dissimilarity between the selected solution and the neighbor solution using equations \ref{eq7} and \ref{eq8}. In equation \ref{eq8}, Jaccard similarity coefficient is calculated. 
\begin{eqnarray}
similiarity(x_i, x_j) &=& \frac{M_{11}}{M_{01} +M_{10}+M_{11}}\label{eq7}\\
dissimiliarity(x_i, x_j) &=& 1 - similiarity(x_i, x_j)\label{eq8}
\end{eqnarray}

where $M_{11}$ is the number of bits where both $X_i$ and $X_n$ have a value 1. $M_{01}$ and $M_{10}$ are determined similarly according to the bits of $X_i$ and $X_n$. The new candidate solution is generated by Equation \ref{eq9} and integer nonlinear programming model \ref{eq10}.

\begin{eqnarray}
 dissimiliarity(v_i, x_i)& \approx &\phi\times dissimiliarity(x_i, x_j)\label{eq9}\\
 \min\:\: | dissimiliarity(v_i, x_i)& -& \phi\times dissimiliarity(x_i, x_j)|\label{eq10}
\end{eqnarray}

\indent \indent \indent \indent  \text{Subject to:}
\begin{align*}
    M_{11} + M_{01} = n_1\\
    M_{10} \leq n_0\\
    \{M_{10},M_{11}, M_{01}\} \geq 0 \:\text{and}\: \in \mathbb{Z}
\end{align*}
$\phi$ is the random positive scaling factor. The minimum possible value is determined according to the difference between the candidate solution and the selected solution. Since it is not possible to provide equality in all conditions, there is the approximately equal expression '$\approx$'.  To solve Equation (9), it is necessary to solve Equation (10) first by using integer programming techniques. In the constraints, $n_1$ and $n_0$ represent the number of 1 bit and 0 bit in the selected solution, respectively. A new candidate solution is generated by using the bits of $M_{11}$,$M_{01}$ and $M_{10}$ obtained by solving model (10).  For more information and examples, the reader can refer to \cite{ref17}.

\subsection{ABCbin}
In this ABC variant proposed by Kıran, continuous decision variables are converted into binary vector by Equation \ref{eq11}. In the ABCbin algorithm, Equation \ref{eq4} is used as the neighborhood operator. Since it is not possible to apply decision variables in continuous space to the problem, the binary vector ($z_i$) obtained by Equation \ref{eq11} is used.

\begin{equation}
    \label{eq11}
    z_i = round(|x_i\textbf{mod2}|)\textbf{mod2}
\end{equation}

The original ABC algorithm can be easily adapted to binary problems by this variant. The ABCbin algorithm was implemented for different population numbers and competitive results against the binABC and disABC algorithms were obtained \cite{ref20}.

\section{Improved Binary Artificial Bee Colony (ibinABC)}
The ABC algorithm includes updating the value of only one decision variable over D decision variables in each update process. However, new generation swarm intelligence methods such as WOA and GWO update all decision variables in each iteration. While updating a single decision variable strengthens the exploitation phase, it weakens the exploration phase and raises the problem of getting stuck in a local minimum \cite{ref21}. 

Two basic arrangements have been conducted in the proposed binary variant of the ABC algorithm. The first is to increase the convergence rate and strengthen the exploration phase by considering a variable number of bits for the neighborhood operator. Instead of using a fixed value in each iteration, this value is determined by equation \ref{eq12} based on the iteration.

\begin{equation}\label{eq12}
    d_t = rand(0,\alpha)+e^{-(\frac{t}{t_{max}})\times 0.1\times D}+1	
\end{equation} 
$\alpha$  is the perturbation coefficient and it is a random integer variable used to prevent exponential decrease. $d_t$ is number of updated bits, PD refers to the problem dimension, $t$ and $t_{max}$  are the current iteration number and the number of maximum iterations, respectively. $d_t$ tends to decrease in each iteration. Therefore, in the first iterations, the exploration phase is strengthened by updating more bits of more selected solutions while a candidate solution is generated. The exploitation phase is strengthened by reducing the number of bits processed when approaching the value of $t_{max}$. At this stage, it is aimed to make changes in more than one bit by using a perturbation number and to prevent the problem of getting stuck in a local minimum.  Since $d_t$ decreases over the course of iterations, the more decision variables are updated at the initial stage. Towards the end of iterations, the less decision variables are updated. This modification balances trade-off between exploration-exploitation phases.

The proposed method employs the updated version of Equation \ref{eq6} proposed by Kıran et al. as a neighborhood operator. It utilizes a neighbor solution or selected solution randomly when $\vartheta$ parameter in Equation \ref{eq6} is selected as 0.5. It does not make any difference if the neighbor solution is better or worse than the selected solution. This approach weakens the exploitation phase because it involves a more random process.

In our proposed variant, parameter $\vartheta$ is adaptively determined by Equation \ref{eq13}. If the neighbor solution is better than the selected solution, then State 2 occurs by setting the value of  $\vartheta$ to 0. Otherwise, the value of  $\vartheta$ is determined depending on the iterations. In the method, the value of  $\vartheta$ decreases linearly with the iterations. Thus, a worse solution at the beginning of the search is also allowed. 
\begin{equation}\label{eq13}
    \vartheta = \begin{cases} 
        Q_{max}-(\frac{Q_{max}-Q_{min}}{t_{max}}) \times t & Fit(x_n) < Fit(x_i)\\
        0 & \text{otherwise}
        \end{cases}
\end{equation}
where $Q_{start}$ and $Q_{end}$ are the upper and lower bound of a particular range, respectively.  If the neighbor solution is better than current solution $\vartheta$ assigned as zero and selected bit of candidate solution copied from neighborhood solution. Otherwise there is a low probability to copied from neighborhood solution.

The ibinABC is created adding the two major modification to binABC. The pseudocode of the proposed method is given in Algorithm \ref{pseudo}.

\begin{algorithm}[h]

	\caption{The pseudo code of the ibinABC}
	\label{pseudo}
	\begin{algorithmic}[1]
	    \State Set the parameters ($N$, $t_{max}$, $Q_{start}$, $Q_{end}$, $\alpha$, $limit$)
	    \State Generate initial population
	    \State Memorize the best solution
		 \While{termination criteria is not met}
		 \For{each employed bee}
		 \State Select neighbor using Roulette-Wheel selection
		 \State Determine $\vartheta$ using eq.13
		 \State Determine $d_t$ using eq.12
	     \For{i=1 to $d_t$}
	      \State Apply neighborhood operator using eq.6
	      \EndFor 
	      \State Evaluate the candidate solution
	      \If{the candidate solution is better than the selected solution}
	      \State Set the current solution as the candidate solution and reset the trial
	      \Else
	      \State Increment the trial
	      \EndIf
	      \EndFor
	      \For{each onlooker bee according to the probability}
		 \State Select neighbor using Roulette-Wheel selection
		 \State Determine $\vartheta$ using eq.13
		 \State Determine $d_t$ using eq.12
	     \For{i=1 to $d_t$}
	      \State Apply neighborhood operator using eq.6
	      \EndFor 
	      \State Evaluate the candidate solution
	      \If{the candidate solution is better than the selected solution}
	      \State Set the current solution as the candidate solution and reset the trial
	      \Else
	      \State Increment the trial
	      \EndIf
	      \EndFor
	      
	      \If{there exists food source exceeding limit, then choose one}
	      \State Replace it with the new solution and reset the trial
	      \EndIf
	      
		 \State Memorize
		 \EndWhile
	\end{algorithmic}
\end{algorithm}
\section{Uncapacitated Facility Location Problem}
In this section, we introduce a binary optimization problem named the Uncapacitated Facility Location Problem (UFLP) used to show the effectiveness of the proposed variant. The UFLP aims to find the locations of customers whose demands are previously determined and the locations where potential facilities can be built. Each facility has a set-up cost and the transportation cost between the customer and the facility. The main purpose of the problem is to locate the facilities to be built with a minimum total cost and to determine the location of the facilities used by the customers. The problem is called uncapacited because the facilities are assumed to have the service capacity to meet all the customers' demands.

To define the UFLP mathematically, let m and n be the number of potential facilities to be built and the number of customers, respectively. Let S be the shipment cost matrix and  $S_{i,j}$ be the transportation cost between facility $i$ and customer $j$. Let $F$ be the set-up cost vector. $F_i$ represents the initial installation cost of facility $i$. $y_{i,j}$ and $x_i$ are binary decision variables defined as follows:
\begin{equation}
\label{eq14}
    y_{i,j}=\begin{cases}
			1, & \text{if customer j is served by facility i}\\
            0, & \text{otherwise}
		 \end{cases}
\end{equation}
and
\begin{equation}
\label{eq15}
    x_{i,j}=\begin{cases}
			1, & \text{if facility i is built}\\
            0, & \text{otherwise}
		 \end{cases}
\end{equation}
The objective function is defined in Equation \ref{eq:obj}, \ref{eq:cons_1}).
\begin{equation}
\label{eq:obj}
    \min{f = \sum_{i=1}^n{\sum_{j=1}^m}{c_{i,j} y_{i,j}} + \sum_{i=1}^n{f_i x_i} }
\end{equation}
Subject to:

\begin{equation}
\label{eq:cons_1}  
    \begin{split}
        &\sum_{i=1}^n{y_{i,j}=1}, \quad j=1,2,...,m  \\
        &\quad \quad y_{i,j} \geq x_i, \quad i=1,2,...,n  \quad j=1,2,...,m \\ %\label{eq:cons_2} \label{eq:cons_3}
        &\quad \quad y_{i,j}, x_{i} \in \{ 0,1 \} 
\end{split}
\end{equation}

The vector $x$ used as the decision variable in the UFLP determines whether the facilities are built or not.  If a potential facility is built at its location, then the corresponding decision variable takes the value 1, otherwise, it takes the value 0. Since all decision variables of the problem are binary, the UFLP belongs to the class of binary integer programming problems.

OR-Library is a collection of problem instances for a variety of combinatorial optimization problems. Some problem instances including the number of facility locations and the number of customers in OR-Library are given in Table \ref{table:orlib} \cite{ref23}.

\begin{table}
\label{table:orlib}
\caption{OR-Lib UFLP Dataset description}
\centering
\begin{tabular}{lll} 
\hline
Problem
  instance & Problem
  size                         & Optimal
  cost value        \\ 
\hline
Cap71              &  16 $\times$ 50  &  {932,615.75}        \\
Cap72              &  16 $\times$  50                    &  {977,799.40}        \\
Cap73              &  16 $\times$  50                    &  {1,010,641.45}      \\
Cap74              &  16 $\times$ 50                    &  {1,034,976.98}      \\
Cap101             &  25 $\times$  50                    &  {796,648.44}        \\
Cap102             &  25 $\times$ 50                    &  {854,704.20}        \\
Cap103             &  25 $\times$  50                    &  {893,782.11}        \\
Cap104             &  25 $\times$ 50                    &  {928,941.75}        \\
Cap131             &  50 $\times$ 50                    &  {793,439.56}        \\
Cap132             &  50 $\times$ 50                    &  {851,495.33}        \\
Cap133             &  50 $\times$ 50                    &  {893,076.71}        \\
Cap134             &  50 $\times$ 50                    &  {928,941.75}        \\
CapA               &  100 $\times$ 1000                 &  {17,156,454.48}     \\
CapB               &  100 $\times$ 1000                 &  {12,979,071.58}     \\
CapC               &  100 $\times$ 1000                 &  {11,505,594.33}    
\end{tabular}
\end{table}
\section{Experimental Results}
In this section, we discuss the success of the proposed binary ABC algorithm for solving the UFLP and give an extensive comparison with the other methods existing in the literature for the same problem.
\subsection{Parameter Tuning}
We first conducted a parameter tuning process for the experimental study. To determine the best algorithm parameters, we tested several values of $Q_{start}, Q_{end}, limit$ and the number of individuals in the population ($N$). Accordingly, we created 24 implementations consisting of three permutations of $Q_{start}$ and $Q_{end}$  ($Q_{start}=0.5;Q_{end}=0.3$ and $Q_{start}=0.5;Q_{end}=0.1$and $Q_{start}=0.3;Q_{end}=0.1$), four different values of limit  ($N\times D\times 0.5, N\times D,N \times D\times 2$ and $N\times D\times 4$) and two values of $N=(20,40)$.  We tested these 24 implementations on problem instances CapA, CapB and CapC  for 80000 function evaluations by running 30 times. Other problem instances have not been taken into the test environment because they are easier to solve and the results are not distinctive.
Table 3 shows a summary of the computational results of 24 implementations tested on CapA problem instance. The summary of the results over 30 runs consists of the average minimum cost value,  the worst minimum cost value, the best minimum cost value, the standard deviation of the cost values, and the number of hits with optimal values over 30 runs. Gap represents the average gap between the optimal value and obtained mean value for 30 different runs and is calculated using Equation (18).

\begin{equation}
        Gap=\dfrac{(Mean-Optimum)}{Optimum} \times 100
\end{equation}

As can be seen in Table 3, the optimal value for 10 implementations was achieved in 30 runs, that is, in each run. These optimal values were obtained  four times for N=20  and six times for N=40. However, the optimal value was achieved 2,1,3,4 times, respectively, for different limit values in Table 3. The optimal value was reached 2, 2, 6 times, respectively, for different Q values in Table 3.
Table 3. Parameter tuning on problem instance CapA.

\begin{table}[ht!]
\caption{Parameter tuning for CapA instance}
\centering
\small
\begin{tabular}{cccccccc}
\hline
~                                         & ~    & \multicolumn{3}{c}{N:20}                      & \multicolumn{3}{c}{N:40}                       \\ 
\cline{3-8}
Limit                                         & ~    & Q=\{0.5,0.3\} & Q=\{0.5,0.1\} & Q=\{0.3,0.1\} & Q=\{0.5,0.3\} & Q=\{0.5,0.1\} & Q=\{0.3,0.1\}  \\ 
\hline
\multirow{4}{*}{ $N\times D \times 0.5$} & Mean & 17211513.98   & 17211416.95   & 17202659.73   & 17156454.48   & 17157257.31   & 17156454.48    \\
                                          & Std  & 118060.08     & 121645.08     & 124472.71     & 0.00          & 4397.31       & 0.00           \\
                                          & Gap  & 0.321         & 0.320         & 0.269         & 0.000         & 0.005         & 0.000          \\
                                          & Hit  & 23            & 22            & 23            & 30            & 29            & 30             \\ 
\hline
\multirow{4}{*}{{$N\times D$}}   & Mean & 17158862.99   & 17177089.92   & 17162797.73   & 17157257.31   & 17163600.57   & 17156454.48    \\
                                          & Std  & 7349.05       & 57841.98      & 34743.44      & 4397.31       & 34869.86      & 0.00           \\
                                          & Gap  & 0.014         & 0.120         & 0.037         & 0.005         & 0.042         & 0.000          \\
                                          & Hit  & 27            & 25            & 29            & 29            & 28            & 30             \\ 
\hline
\multirow{4}{*}{{ $N\times D \times 2$}} & Mean & 17157257.31   & 17156454.48   & 17156454.48   & 17157257.31   & 17157257.31   & 17156454.48    \\
                                          & Std  & 4397.31       & 0.00          & 0.00          & 4397.31       & 4397.31       & 0.00           \\
                                          & Gap  & 0.005         & 0.000         & 0.000         & 0.005         & 0.005         & 0.000          \\
                                          & Hit  & 29            & 30            & 30            & 29            & 29            & 30             \\ 
\hline
\multirow{4}{*}{{  $N\times D \times 4$}} & Mean & 17158862.99   & 17156454.48   & 17156454.48   & 17156454.48   & 17162797.73   & 17156454.48    \\
                                          & Std  & 7349.05       & 0.00          & 0.00          & 0.00          & 34743.44      & 0.00           \\
                                          & Gap  & 0.014         & 0.000         & 0.000         & 0.000         & 0.037         & 0.000          \\
                                          & Hit  & 27            & 30            & 30            & 30            & 29            & 30             \\
\hline
\end{tabular}
\end{table}

\begin{table}
\caption{Parameter tuning for CapB instance}
\centering
\begin{tabular}{cccccccc}
~                               & ~    & \multicolumn{3}{c}{N:20}                      & \multicolumn{3}{c}{N:40}                       \\ 
\cline{3-8}
~ Limit                              & ~    & Q=\{0.5,0.3\} & Q=\{0.5,0.1\} & Q=\{0.3,0.1\} & Q=\{0.5,0.3\} & Q=\{0.5,0.1\} & Q=\{0.3,0.1\}  \\ 
\hline
\multirow{4}{*}{$N\times D \times 0.5$} & Mean & 13040252.87   & 13026593.82   & 13015214.76   & 13000993.81   & 12997280.3    & 12985106.11    \\
                                & Std  & 54956.31      & 61875.43      & 45090.86      & 33861.62      & 31331.36      & 15256.78       \\
                                & Gap  & 0.471         & 0.366         & 0.278         & 0.169         & 0.140         & 0.046          \\
                                & Hit  & 9             & 14            & 12            & 18            & 17            & 22             \\ 
\hline
\multirow{4}{*}{$N\times D$}   & Mean & 13017483      & 13014189.49   & 12995042.72   & 12993104.83   & 13000429.62   & 12986432.62    \\
                                & Std  & 47236.15      & 43546.89      & 38948.36      & 32061.70      & 33164.37      & 20707.79       \\
                                & Gap  & 0.296         & 0.271         & 0.123         & 0.108         & 0.165         & 0.057          \\
                                & Hit  & 15            & 15            & 22            & 23            & 18            & 25             \\ 
\hline
\multirow{4}{*}{$N\times D \times 2$} & Mean & 12996226.12   & 12992661.06   & 12988144.53   & 12993190.01   & 12992449.05   & 12988220.02    \\
                                & Std  & 31866.51      & 28794.20      & 23762.93      & 26057.15      & 28303.10      & 20970.35       \\
                                & Gap  & 0.132         & 0.105         & 0.070         & 0.109         & 0.103         & 0.070          \\
                                & Hit  & 20            & 20            & 24            & 20            & 21            & 22             \\ 
\hline
\multirow{4}{*}{$N\times D \times 4$} & Mean & 12989864.89   & 12997800.6    & 12998207.46   & 12992871.59   & 13001931.86   & 12987621.8     \\
                                & Std  & 24533.19      & 33722.88      & 32646.16      & 27174.68      & 34542.24      & 23511.75       \\
                                & Gap  & 0.083         & 0.144         & 0.147         & 0.106         & 0.176         & 0.066          \\
                                & Hit  & 22            & 17            & 21            & 21            & 16            & 25             \\
\hline
\end{tabular}
\end{table}

The results for problem instance CapB after 30 runs are presented in Table 4.  According to the table, the best solutions are obtained for  N=40 and $Q_{start}=0.3; Q_{end}=0.1$. When the limit parameter is analyzed, the best results are obtained when $Limit=N \times D$ and $Limit=N \times  D \times 4$.  24 hits, i.e., the number of runs with the optimal value, are achieved for $N=20$, $Q_{start}=0.3; Q_{end}=0.1$ and $Limit=N\times D \times 2$. 
	
The results for problem instance CapC after 30 runs are given in Table 5.  As can be seen in the table, the optimal value is achieved in 9 runs among 30 runs when $N=40$, $Limit=N\times D\times 2$ and  $Limit=N\times D\times 0.5$.  13 hits are achieved  for $N=20$, $Q_{start}=0.3; Q_{end}=0.1$ and $Limit=N\times D\times 2$. 

As can be seen in Tables 1 to 3, the best parameter values were determined as $N=20$, $Q_{start}=0.3; Q_{end}=0.1$ and $Limit=N\times D\times 2$. We used the computational results obtained using these parameters for comparison with the other algorithms.

\begin{figure}[h!]
    \centering
    \includegraphics[scale=0.3]{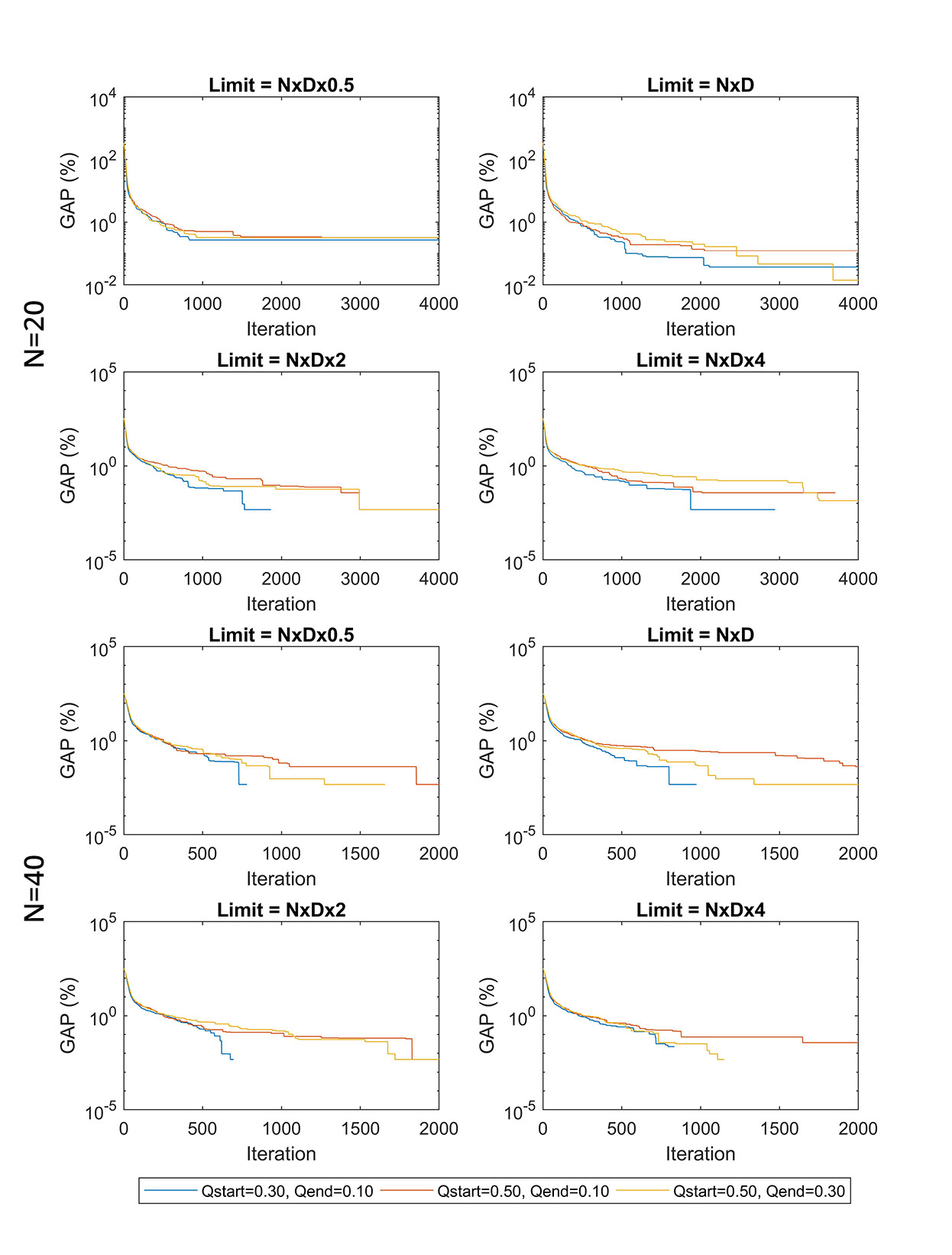}
    \caption{Parameter tuning on problem instance CapA.}
    \label{capaconvergence}
\end{figure}

\begin{table}
\caption{Parameter tuning for CapC instance}
\centering
\begin{tabular}{cccccccc}
~                               & ~    & \multicolumn{3}{c}{N:20}                      & \multicolumn{3}{c}{N:40}                       \\ 
\cline{3-8}
~ Limit                              & ~    & Q=\{0.5,0.3\} & Q=\{0.5,0.1\} & Q=\{0.3,0.1\} & Q=\{0.5,0.3\} & Q=\{0.5,0.1\} & Q=\{0.3,0.1\}  \\ 
\hline
\multirow{4}{*}{$N\times D \times 0.5$} & Mean & 11537949.08   & 11531590.6    & 11526389.19   & 11519621.62   & 11524032.61   & 11512198.34    \\
                                & Std  & 36722.90      & 27329.10      & 29639.45      & 16432.87      & 22406.63      & 9798.03        \\
                                & Gap  & 0.281         & 0.226         & 0.181         & 0.122         & 0.160         & 0.057          \\
                                & Hit  & 1             & 4             & 7             & 4             & 6             & 9              \\ 
\hline
\multirow{4}{*}{$N\times D$}   & Mean & 11524585.71   & 11522732.29   & 11515450.4    & 11515816.97   & 11521586.29   & 11515181.83    \\
                                & Std  & 24926.34      & 16770.08      & 10456.57      & 12328.59      & 17861.71      & 10868.37       \\
                                & Gap  & 0.165         & 0.149         & 0.086         & 0.089         & 0.139         & 0.083          \\
                                & Hit  & 2             & 4             & 6             & 9             & 5             & 4              \\ 
\hline
\multirow{4}{*}{$N\times D \times 2$} & Mean & 11514048.28   & 11516527.21   & 11512756.39   & 11519689.91   & 11521590.87   & 11514458.39    \\
                                & Std  & 10001.90      & 11814.39      & 11326.02      & 13314.59      & 14937.16      & 11717.41       \\
                                & Gap  & 0.073         & 0.095         & 0.062         & 0.123         & 0.139         & 0.077          \\
                                & Hit  & 4             & 7             & 13            & 6             & 4             & 7              \\ 
\hline
\multirow{4}{*}{$N\times D \times 4$} & Mean & 11516659.14   & 11520120.18   & 11514276.73   & 11523715.66   & 11520523.78   & 11513374.05    \\
                                & Std  & 13096.58      & 15931.59      & 10349.29      & 20569.69      & 15529.49      & 12098.64       \\
                                & Gap  & 0.096         & 0.126         & 0.075         & 0.158         & 0.130         & 0.068          \\
                                & Hit  & 6             & 6             & 6             & 3             & 3             & 8              \\
\hline
\end{tabular}
\end{table}
Fig. \ref{capaconvergence} shows convergence graphs of the results obtained after 30 runs of different implementations for problem instance CapA. Each sub-figure was obtained for different values of  $Q$, limit and population. The first four and the next four figures were obtained for $N=20$ and $N=40$, respectively. For  $N=20$, if the limit value is not selected properly, then there is a problem of late local convergence and falling into a local minimum ($Limit= N\times D \times 0.5$ and $Limit=N \times D$). It is seen that $Q_{start}=0.3$ and $Q_{end}=0.01$ with the other limit values have faster convergence and optimal values are achieved. We can observe that there is a decrease in convergence rate for the other values of Q.  Moreover, as can easily be seen that $Q_{start}=0.3$ and $Q_{end=0.01}$  has faster convergence speed for $N=40$ and all values of limit compared to the other values of $Q$.

\begin{figure}[h!]
    \centering
    \includegraphics[scale=0.3]{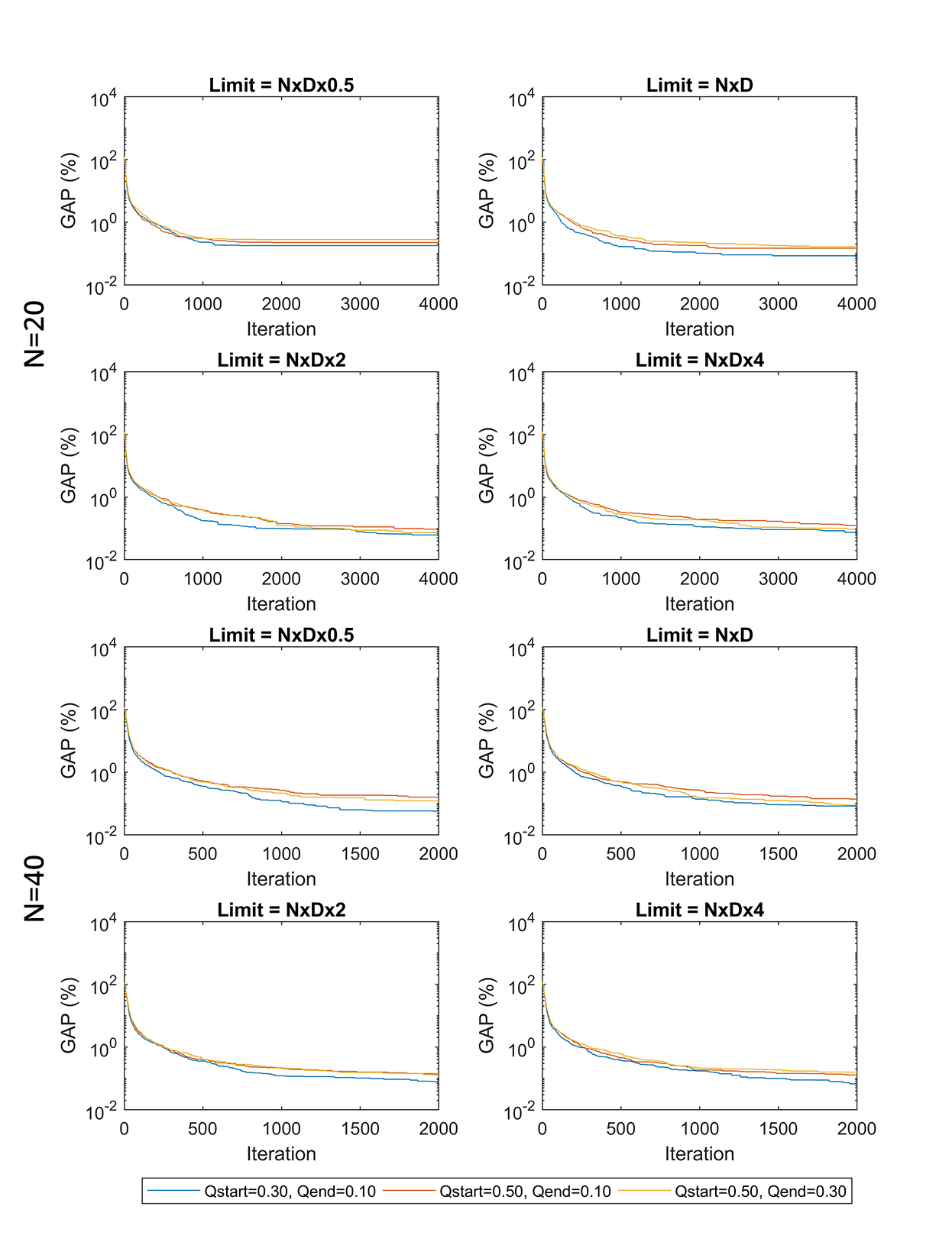}
    \caption{Parameter tuning on problem instance CapC.}
    \label{capcconvergence}
\end{figure}

\subsection{Comparison of methods}
In this section, we compared the methods in the literature that are applied to the UFLP with a certain success with the proposed method. The methods and results to be used for comparison are taken from \cite{ref18} and \cite{ref22}. In the proposed method and other methods for making a fair comparison, the size of the population is set to 40 and the maximum number of iterations is set to 2000.  The parameters used in the methods are presented in Table 6. The proposed method is implemented in C programming language and other methods are implemented in different platforms. Since the method we propose is faster than the methods in the literature, it will not provide a fair comparison. Therefore, we do not present the CPU time.

\begin{table}
\caption{Parameter configuration of ABC variants}
\centering
\begin{tabular}{lllll}
~                       & binABC  & DisABC     & ABCbin  & ibinABC    \\ 
\hline
Population Size         & 40      & 40         & D       & 20         \\
Max number of iteration & 2000    & 2000       & 1000    & 4000       \\
Limit                   & N x D/4 & 2.5 x N xD & N x D/2 & N x D x 2  \\
\hline
\end{tabular}
\end{table}
We present the comparison of the proposed ibinABC algorithm with the other ABC variants in Table 7. According to the table, the proposed method is superior to the other methods for some problem instances.  Our method did not obtain the optimal values among 30 runs only for problem instances CapB and CapC. However, it has a very low GAP value compared to other methods. The method produces very successful results not only for small-size problem instances but also for large-size instances. As can be clearly seen in the table, the proposed method is the best among the other methods we compared.
\begin{table}
\centering
\begin{tabular}{lllllllll}
~      & \multicolumn{2}{l}{binABC} & \multicolumn{2}{l}{DisABC} & \multicolumn{2}{l}{ABCbin} & \multicolumn{2}{l}{ibinABC}  \\ 
\cline{2-9}
~      & Gap      & Std             & Gap  & Std                 & Gap  & Std                 & Gap  & Std                   \\ 
\cline{2-9}
Cap71  & 0.00     & 0.00            & 0.00 & 0.00                & 0.00 & 0.00                & 0.00 & 0.00                  \\
Cap72  & 0.00     & 0.00            & 0.00 & 0.00                & 0.00 & 0.00                & 0.00 & 0.00                  \\
Cap73  & 0.00     & 0.00            & 0.00 & 0.00                & 0.00 & 0.00                & 0.00 & 0.00                  \\
Cap74  & 0.00     & 0.00            & 0.00 & 0.00                & 0.00 & 0.00                & 0.00 & 0.00                  \\
Cap101 & 0.00     & 0.00            & 0.00 & 0.00                & 0.00 & 0.00                & 0.00 & 0.00                  \\
Cap102 & 0.00     & 0.00            & 0.00 & 0.00                & 0.00 & 0.00                & 0.00 & 0.00                  \\
Cap103 & 0.00     & 0.00            & 0.00 & 0.00                & 0.01 & 85.67               & 0.00 & 0.00                  \\
Cap104 & 0.00     & 0.00            & 0.00 & 0.00                & 0.00 & 0.00                & 0.00 & 0.00                  \\
Cap131 & 0.00     & 0.00            & 0.62 & 2,337.64            & 0.20 & 1,065.73            & 0.00 & 0.00                  \\
Cap132 & 0.00     & 0.00            & 0.09 & 813.37              & 0.02 & 213.28              & 0.00 & 0.00                  \\
Cap133 & 1,215.00 & 200.24          & 0.03 & 359.03              & 0.07 & 561.34              & 0.00 & 0.00                  \\
Cap134 & 0.00     & 0.00            & 0.00 & 0.00                & 0.00 & 0.00                & 0.00 & 0.00                  \\
CapA   & 2.96     & 236,833.50      & 0.15 & 74,782.61           & 3.17 & 268,685.20          & 0.00 & 0.00                  \\
CapB   & 2.51     & 9,143.13        & 3.30 & 109,738.50          & 2.82 & 88,452.80           & 0.07 & 23,762.93             \\
CapC   & 2.58     & 82,312.70       & 4.70 & 95,778.78           & 2.04 & 78,162.20           & 0.06 & 11,326.02             \\
\hline
\end{tabular}
\end{table}

We also compared the proposed ibinABC algorithm with some meta-heuristic algorithms (GA, binary Artificial Algae Algorithm (binAAA) and binary PSO) and presented the results in Table 8. As can be seen in the table, the results for problem instances CapA, CapB and CapC are distinctive. The BinAAA and ibinABC algorithms can compete on these three problem instances. The ibinABC achieved more hits for CapB and CapC than the binAAA algorithm and yielded results closer to the optimal value.  

\begin{table}
\centering
\begin{tabular}{lllllllllllll}
~      & \multicolumn{3}{l}{GA-SP} & \multicolumn{3}{l}{BPSO}  & \multicolumn{3}{l}{binAAA} & \multicolumn{3}{l}{ibinABC}  \\ 
\cline{2-13}
~      & Gap   & Std. Dev.  & Hit  & Gap   & Std. Dev.   & Hit & Gap   & Std. Dev.  & Hit   & Gap   & Std. Dev.  & Hit     \\ 
\hline
Cap71  & 0.00  & 0.00       & 30   & 0.000 & 0.000       & 30  & 0.000 & 0.000      & 30    & 0.000 & 0.000      & 30      \\
Cap72  & 0.000 & 0.000      & 30   & 0.000 & 0.000       & 30  & 0.000 & 0.000      & 30    & 0.000 & 0.000      & 30      \\
Cap73  & 0.067 & 899.650    & 19   & 0.024 & 634.625     & 26  & 0.000 & 0.000      & 30    & 0.000 & 0.000      & 30      \\
Cap74  & 0.000 & 0.000      & 30   & 0.009 & 500.272     & 29  & 0.000 & 0.000      & 30    & 0.000 & 0.000      & 30      \\
Cap101 & 0.068 & 421.655    & 11   & 0.043 & 428.658     & 18  & 0.000 & 0.000      & 30    & 0.000 & 0.000      & 30      \\
Cap102 & 0.000 & 0.000      & 30   & 0.010 & 321.588     & 28  & 0.000 & 0.000      & 30    & 0.000 & 0.000      & 30      \\
Cap103 & 0.064 & 505.036    & 6    & 0.049 & 521.237     & 14  & 0.000 & 0.000      & 30    & 0.000 & 0.000      & 30      \\
Cap104 & 0.000 & 0.000      & 30   & 0.041 & 1,432.239   & 28  & 0.000 & 0.000      & 30    & 0.000 & 0.000      & 30      \\
Cap131 & 0.068 & 720.877    & 16   & 0.171 & 1,505.749   & 10  & 0.000 & 0.000      & 30    & 0.000 & 0.000      & 30      \\
Cap132 & 0.000 & 0.000      & 30   & 0.058 & 1,055.238   & 21  & 0.000 & 0.000      & 30    & 0.000 & 0.000      & 30      \\
Cap133 & 0.091 & 685.076    & 10   & 0.083 & 690.192     & 10  & 0.000 & 0.000      & 30    & 0.000 & 0.000      & 30      \\
Cap134 & 0.000 & 0.000      & 30   & 0.195 & 2,594.211   & 18  & 0.000 & 0.000      & 30    & 0.000 & 0.000      & 30      \\
CapA   & 0.046 & 22,451.206 & 24   & 1.691 & 319,855.431 & 8   & 0.000 & 0.000      & 30    & 0.000 & 0.000      & 30      \\
CapB   & 0.584 & 66,658.649 & 9    & 1.403 & 135,326.728 & 5   & 0.248 & 39,224.744 & 15    & 0.070 & 23,762.929 & 24      \\
CapC   & 0.705 & 51,848.280 & 2    & 1.622 & 115,156.444 & 1   & 0.295 & 29,766.311 & 1     & 0.062 & 11,326.015 & 13      \\
\hline
\end{tabular}
\end{table}
\section{Conclusion}
In this paper, we proposed a binary variant of the ABC algorithm in order to successfully apply the ABC algorithm to binary optimization problems. The proposed method aims to increase the convergence rate by updating some decision variables in each iteration and to prevent the problem of getting stuck in a local minimum. Another improvement is the use of adaptive parameters in XOR-based logical operators. We conducted a separate study to determine these parameters experimentally and found the best configuration.  We discussed the success of the proposed binary ABC algorithm for solving the UFLP and presented an extensive comparison with the other methods existing in the literature. According to the computational tests, the proposed method is superior to the other methods. The ibinABC showed the best performance among all problem instances of the UFLP taken from OR-Library.

\end{document}